\documentclass[10pt,twocolumn,letterpaper]{article}
\usepackage[utf8]{inputenc}
\usepackage{cvpr}              

\usepackage{graphicx}
\usepackage{amsmath}
\usepackage{amssymb}
\usepackage{booktabs}
\usepackage{multirow}
\usepackage{bbm}
\usepackage{xcolor}
\usepackage{tabulary}
\usepackage{tabularx}
\usepackage{colortbl}
\usepackage{etoolbox}
\definecolor{mygray}{gray}{0.6}
\definecolor{mygray-bg}{gray}{0.9}
\usepackage{pifont}
\newcommand{\xmark}{\ding{55}}%
\newcommand{\bd}[1]{\textbf{#1}}
\newcommand{\app}{\raise.17ex\hbox{$\scriptstyle\sim$}}

\newcolumntype{x}[1]{>{\centering\arraybackslash}p{#1pt}}

\newlength\savewidth\newcommand\shline{\noalign{\global\savewidth\arrayrulewidth
  \global\arrayrulewidth 1pt}\hline\noalign{\global\arrayrulewidth\savewidth}}
\newcommand{\tablestyle}[2]{\setlength{\tabcolsep}{#1}\renewcommand{\arraystretch}{#2}\centering\footnotesize}

\usepackage[pagebackref,breaklinks,colorlinks]{hyperref}

\usepackage[capitalize]{cleveref}
\crefname{section}{Sec.}{Secs.}
\Crefname{section}{Section}{Sections}
\Crefname{table}{Table}{Tables}
\crefname{table}{Tab.}{Tabs.}

\def\confName{CVPR}
\def\confYear{2022}
\title{TR-MOT: Multi-Object Tracking by Reference}

\author{Mingfei Chen\textsuperscript{\rm 1,3} \quad Yue Liao\textsuperscript{\rm 2} \quad Si Liu\textsuperscript{\rm 2} \quad Fei Wang\textsuperscript{\rm 3}\quad Jenq-Neng Hwang\textsuperscript{\rm 1}\\
\large\textsuperscript{\rm 1} University of Washington \\ \textsuperscript{\rm 2} Institute of Artificial Intelligence, Beihang University \quad\textsuperscript{\rm 3} SenseTime Research}

\begin{document}

\maketitle

\begin{abstract}
Multi-object Tracking~(MOT) generally can be split into two sub-tasks, \emph{i.e.}, detection and association. Many previous methods follow the tracking by detection paradigm, which first obtain detections at each frame and then associate them between adjacent frames. Though with an impressive performance by utilizing a strong detector, it will degrade their detection and association performance under scenes with many occlusions and large motion if not using temporal information.
In this paper, we propose a novel Reference Search~(RS) module to provide a more reliable association based on the deformable transformer structure, which is natural to learn the feature alignment for each object among frames. RS takes previous detected results as references to aggregate the corresponding features from the combined features of the adjacent frames and makes a one-to-one track state prediction for each reference in parallel. Therefore, RS can attain a reliable association coping with unexpected motions by leveraging visual temporal features while maintaining the strong detection performance by decoupling from the detector. Our RS module can also be compatible with the structure of the other tracking by detection frameworks. Furthermore, we propose a joint training strategy and an effective matching pipeline for our online MOT framework with the RS module. Our method 
achieves competitive results on MOT17 and MOT20 datasets.
\end{abstract}

\section{Introduction} \label{intro}
Multi-object Tracking~(MOT), an essential and practical task in video analysis,  benefits many applications, \emph{e.g.}, autonomous driving and smart city. MOT aims to estimate the trajectory of objects with the same identity based on a video sequence that contains multiple objects.
MOT generally can be split into two sub-tasks, \emph{i.e.}, detection and association. However, these two tasks are very different, where one is location-aware, and the other is identity-aware. Therefore, one challenge for MOT is how to extract suitable features in a general framework for these two tasks, the focus of which is very different in feature learning. Another challenge is how to associate two objects with the unpredictable motion of the objects or cameras across two frames.

\begin{figure}[t]
  \begin{center}
  \includegraphics[width=0.95\linewidth]{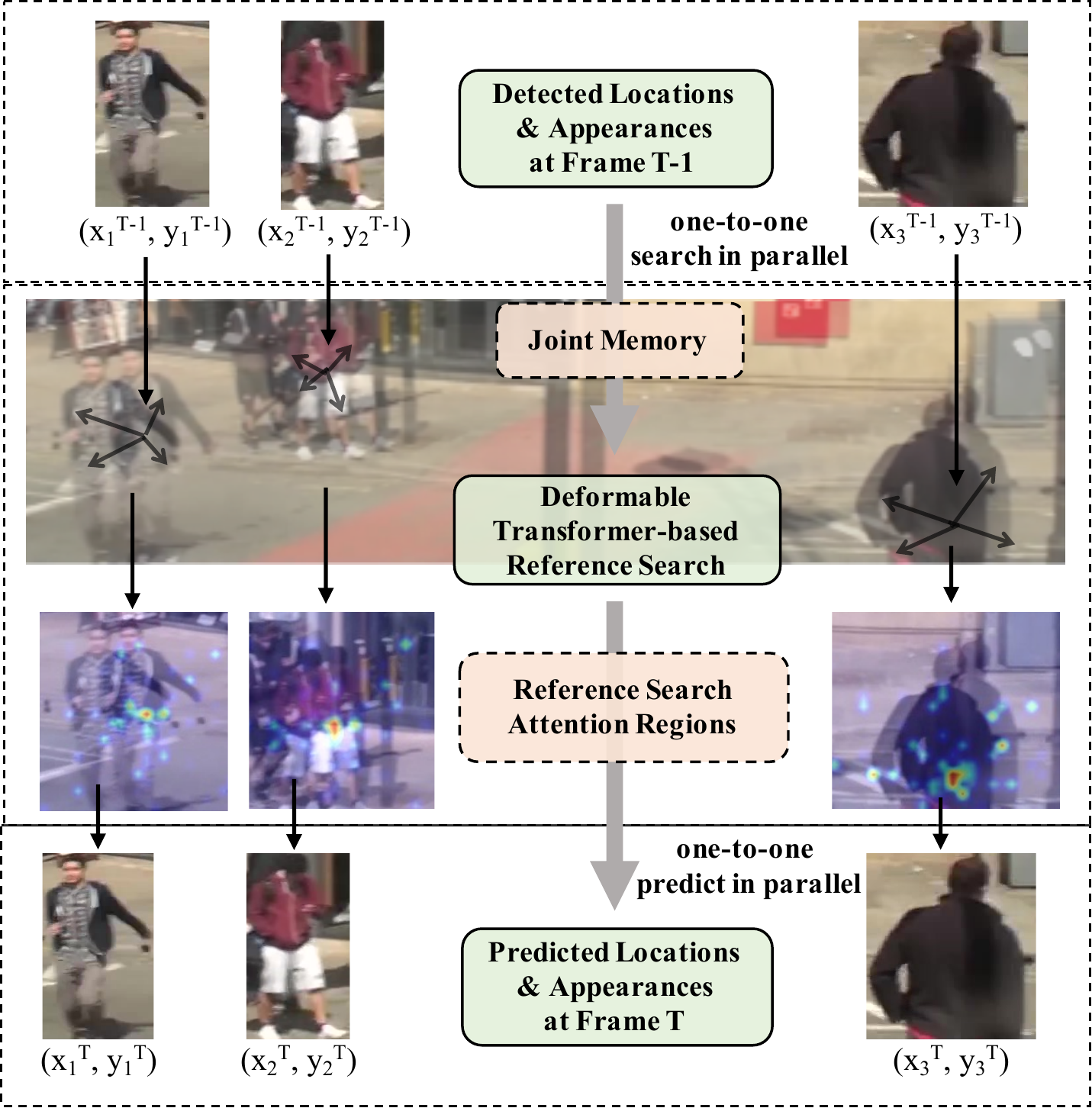}
  \end{center}\vspace{-4mm}
  \caption{In Reference Search~(RS), track locations and appearances at the previous frame are utilized as references to aggregate corresponding features from joint memory (feature summation of adjacent frames), and then predict the track states at the current frame one-to-one in parallel for association. RS can cope with appearance changes and erratic motions by leveraging temporal information and focusing on feature alignments among frames.}
  \label{fig:intro}
\end{figure}

Conventional methods for MOT are mainly following the {\em tracking by detection}~\cite{wojke2017simple, sun2019deep, liang2020rethinking} and {\em joint tracking and detection}~\cite{pang2020tubetk, peng2020chained} paradigms. 
Most previous {\em tracking by detection} methods first detect all objects and then associate them with previous tracks based on the appearance similarity and location overlap. If without learnt association based on visual temporal information, it will degrade their detection and association performance under scenes with many occlusions or large motions. {\em Joint detection and tracking} methods jointly learn detection and association, aggregate features from several adjacent frames to simultaneously predict locations of the associated objects with the same identity at these frames. However, the strong coupling structure of detection and association is easily to overlook the different focuses of these two tasks. 

In this paper, we propose a deformable transformer-based Reference Search~(RS) method to achieve a more reliable association while maintaining the strong detection performance in the tracking by detection paradigm. For tracking at frame $T$, we perform detection only based on the features from frame $T$, and further associate the new detections to the existing tracks, whose current states, including the locations and appearance embeddings, will then be updated from combined features of the frame $T$ and frame $T-1$ by using RS. Specifically, as Fig \ref{fig:intro}, in the RS module, track locations at the previous frame are utilized as the reference point set to aggregate the corresponding features with deformable co-attention from the two adjacent frames. Correspondingly, we introduce a novel joint training strategy and an effective matching pipeline for our framework with the RS module.

We argue that our TR-MOT has the following characteristics and advantages
comparing to previous methods. We first discuss the MOT methods, which utilize visual temporal information to predict association based on a CNN structure such as CenterTrack~\cite{zhou2020tracking}, ChainedTracker~\cite{peng2020chained} and TubeTK~\cite{pang2020tubetk}.  Unlike these previous methods, considering locations of other tracks to have a minor effect on the state prediction for this track, we leverage a more focused set-to-set transformer structure, which is more natural to make a one-to-one state prediction for each previous track in parallel. Moreover, our track state prediction is decoupled from detection and thus less limited by detection-related features. We then distinguish our TR-MOT from some recent transformer-based MOT methods~\cite{Sun2020TransTrackMT, meinhardt2021trackformer,xutranscenter2021, zeng2021motr}. Such transformer-based trackers utilize the detection-related object features in the transformer-based detector to associate objects in other frames. Though attaining impressive detection accuracy, this manner is less reliable for association, because the learnt object features are highly specialized for the detection task, and thus may degrade the association ability. Decoupling from detection, our RS module only takes detection results as location references to learn the feature alignments based on visual temporal information. Without extra relationship modelling, our RS module can also be compatible with other tracking by detection framework structures, predicting new location and appearance embedding of each previously detected track more reliably for the later association.

Our contributions can be mainly summarized as three aspects: 1) We design an effective Reference Search (RS) module for the association, which utilizes the previous track states as references to make a parallel one-to-one state propagation for each existing track based on visual temporal features. 2) We propose an online MOT framework~(TR-MOT) that decouples the detection and association process, including a reliable matching pipeline compatible with RS and the corresponding joint training strategy. 3) Our TR-MOT achieves competitive performance on the MOT dataset, and ablation studies verify the effectiveness of our main designs.

\section{Related Work}
\noindent\textbf{Tracking by detection.}
The methods in the tracking by detection category~\cite{wojke2017simple, sun2019deep, zhang2020fairmot, liang2020rethinking, wang2019towards, zhou2018online, mahmoudi2019multi, yu2016poi} break the MOT task into two steps: detection and association. They first detect all the objects from a single frame, where most of the methods also extract the corresponding appearance feature for each detected object for the later association. Early methods like~\cite{yu2016poi, sun2019deep} deploy an extra appearance extraction network to obtain the representative object appearances. Recent methods such as~\cite{wang2019towards, zhang2020fairmot, liang2020rethinking} simply predict the appearance embedding along with the detection results from the same detector network. 

For association, DeepSORT~\cite{wojke2017simple} proposes an association method that jointly utilizes the location overlaps and appearance affinity to obtain the matching results with the bipartite matching algorithms such as Hungarian Algorithm~\cite{hungarian}. Many current methods such as~\cite{wang2019towards, zhang2020fairmot, liang2020rethinking} also apply such association method, using the probabilistic state estimation methods like Kalman Filter~\cite{Kalman1960ANA} to update the track state, which is hard to handle unexpected motions in a video. There are other methods taking different association strategies. \cite{DBLP:journals/corr/abs-1801-09646} predicts the optical flow to merge the blobs, \cite{braso2020learning} uses graph to connect object nodes from different frames for association, ~\cite{bergmann2019tracking} utilizes the previous detection results as region proposals to detect their associated objects at the current frame. These methods always require carefully designed rules to formulate the MOT association problem into their proposals to guarantee the effectiveness.

\begin{figure*}[htb]
    \vspace{-1mm}
  \centering
\includegraphics[width=1.0\linewidth]{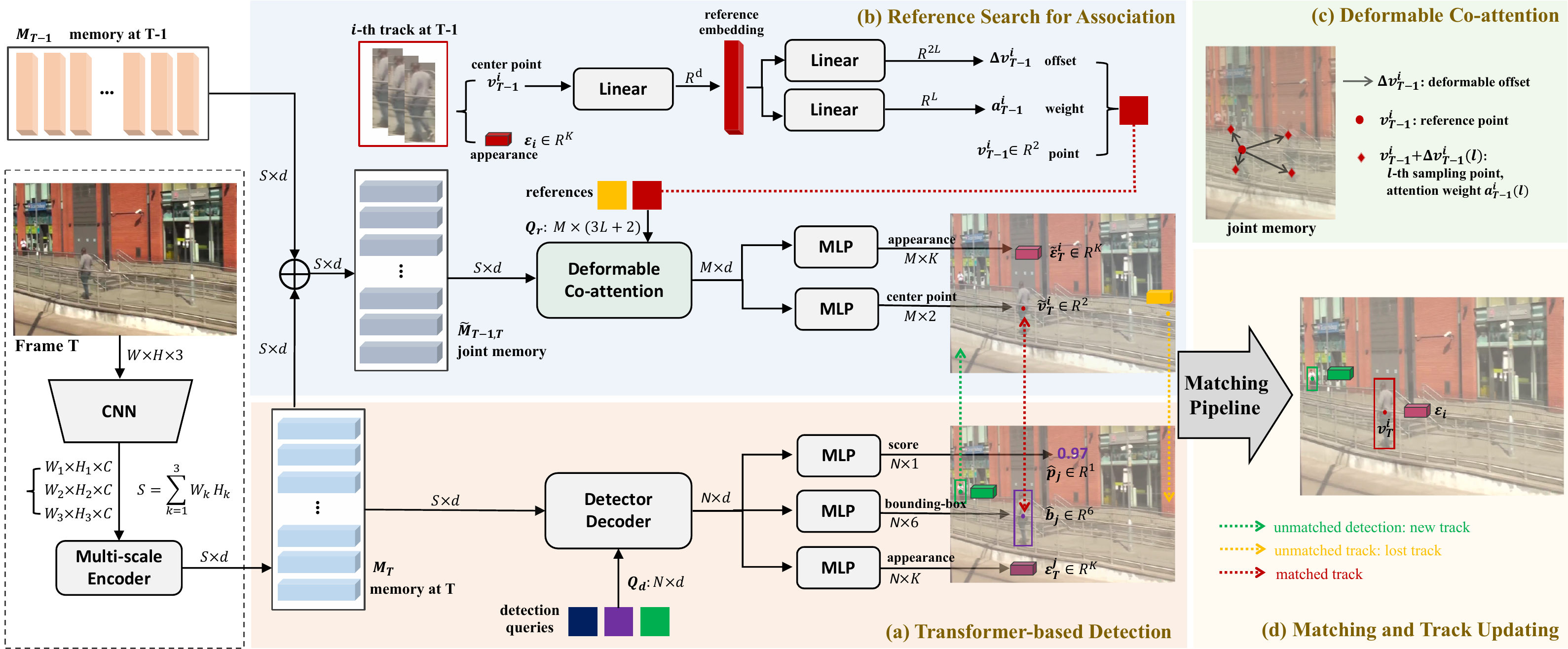}
  \caption{Overview of the proposed framework. (a): For tracking at frame $T$, we perform detection based on the features from frame $T$ to obtain the object appearance embedding and the detection results. (b): The Reference Search~(RS) module is deployed to predict states at frame $T$ for the existing tracks at $T-1$. The RS module takes the feature summation~(joint memory) of frame $T$ and $T-1$ as input. (c): The previous track locations are utilized as reference points to aggregate the corresponding features with deformable co-attention and make a one-to-one prediction of the state~(appearance and location) for each track reference. (d): The predicted track states are associated with the new detections through a matching pipeline. Based on the matching results, previous tracks are finally updated at the frame $T$.}
  \label{framework}\vspace{-2mm}
\end{figure*}

\noindent\textbf{Joint detection and tracking.}
The joint detection and tracking methods~\cite{pang2020tubetk, peng2020chained, zhou2020tracking} predict the association, along with the corresponding detections from adjacent frames. TubeTK~\cite{pang2020tubetk} and ChainedTracker~\cite{peng2020chained} predict the object locations with the same identity in the adjacent frames simultaneously. CenterTrack~\cite{zhou2020tracking} estimates the detection results along with an offset vector, which indicates the motion information from the current object center to its center at the previous frame. Though with a simpler pipeline, such a joint detection and tracking paradigm overlooks the different focus of detection and association. Without a more explicit and reliable feature alignment mechanism for the object in adjacent frames as ~\cite{zhu2017flow} in video object detection, the object features in more than one frame might disturb the detection on the single frame. 

\noindent\textbf{Transformer in multi-object tracking.}
Following the recently popular transformer-based detectors~\cite{carion2020endtoend, zhu2020deformable}, transformer is applied to solve the multi-object tracking problem. Some new tracking methods as~\cite{Sun2020TransTrackMT, meinhardt2021trackformer, zeng2021motr} utilize object features from the previous frame to generate detection queries for the current frame. \cite{xutranscenter2021} uses the transformer with dense queries for tracking the target centers. Though the transformer has impressive detection performance and structure natural for the parallel association, the learned queries for transformer are highly specialized, which are hard to generalize on different tasks directly. Other methods like \cite{pengtransmot2021} designs graph transformers to model the spatial and temporal interactions among objects. However, it requires relationship modelling on each potentially associated node and thus costs more computation and memory.

\vspace{-2mm}\section{TR-MOT} \label{sec:method}\vspace{-1mm}
In this section, we introduce the Tracking-by-Reference MOT~(TR-MOT), an online tracking framework as Fig \ref{framework}, which decouples the detection and association process in multi-object tracking. Precisely, for detection, we deploy a transformer-based detector; for the association, we design the Reference Search~(RS) module to predict the current state of each previous track by leveraging visual temporal information. We propose a corresponding matching pipeline to match the predicted states with newly detected objects and further update the tracks. Finally, a joint training strategy will be introduced to train the whole framework.

\vspace{-2mm}\subsection{Transformer-based Detection} \label{detection} \vspace{-1mm}
Considering the impressive detection performance and set-to-set structure similar to our RS module, we apply the efficient and fast-converging transformer-based detector, called Deformable DETR~\cite{zhu2020deformable} for detection as Fig \ref{framework}~(a). The detector contains an encoder and a decoder. Given the image at frame $T$, after the features are extracted by the backbone CNN, the detector encoder performs multi-scale deformable self-attention on each feature location to aggregate the feature memory $M_T$. Through the linear projections via $N$ learnable queries $Q_d$, we obtain $N$ reference points with their corresponding sampling offsets and attention weights respectively for the multi-scale deformable co-attention in the detector decoder. The decoder aggregates the corresponding features from $M_T$ and decodes $N$ object features for each query one-to-one. From each object feature, three separate heads based on multi-layer perceptions~(MLPs) are used to predict the object bounding-box, category score and appearance embedding, respectively.

Unlike the original deformable DETR, our detector needs to detect the objects even when their center points are outside the image. Thus, we modify deformable DETR for MOT. We clamp each center point into the image and predict the distances from the new center point to four bounding-box sides, respectively, instead of simply predicting the width and the height.
For the appearance embedding prediction, we attach a multi-class classifier~($\varphi$) to the embedding to predict the identity in the training since the ground-truth identity is known. In this way, we can pull the appearance embedding with the same identity altogether, making the embedding more discriminative and representative.

\noindent\textbf{Set-based training.} Considering the loss function should be invariant by a permutation of the prediction set, the Hungarian algorithm~\cite{hungarian} is adopted to assign the ground-truth for supervision. Specifically, we find a one-to-one bipartite matching that minimizes the bounding-box overlap and category difference between the ground-truth $y$ and the $N$ detected objects $\hat y$. 
Denoting the predicted bounding-box, category scores and appearance embedding as $\hat{b}_{{\sigma}}$, $\hat{p}_{{\sigma}}$ and $\hat \varepsilon_{{\sigma}}$ respectively, the detection loss $\mathcal{L}_{\operatorname {det}}$ for the $i$-th ground-truth object with an identity of $\alpha$ is defined as:
\begin{equation}
\small
\vspace{-2mm}
\hspace{-1.5mm}
\begin{split}\mathcal{L}_{\operatorname {det}(i)}= & \mathbbm{1}_{\{c_{i} \neq \varnothing\}}[\lambda_{box}\mathcal{L}_{\operatorname {box}}(b_{i}, \hat{b}_{{\sigma}(i)})+\lambda_{id}\mathcal{L}_{\operatorname {id}}(\alpha, \varphi(\hat \varepsilon_{{\sigma}(i)})] \\ & + \lambda_{cls}\mathcal{L}_{\operatorname {cls}}(c_{i}, \hat{p}_{{\sigma}(i)}),
\end{split}
\nonumber
\vspace{-2mm}
\end{equation}

where $b_i$ and $c_i$ denote the bounding-box and category of the $i$-th ground-truth respectively. The loss $\mathcal{L}_{\operatorname {box}}$~\cite{zhu2020deformable} is composed of a $l_1$ loss and a GIoU loss, both the loss $\mathcal{L}_{\operatorname {cls}}$ and id loss $\mathcal{L}_{\operatorname {id}}$ are Focal Loss~\cite{lin2017focal}. 

\subsection{Reference Search for Association}
We design a Reference Search~(RS) module as Fig \ref{framework}~(b) to predict the current state of previous tracks for the association, using the previous states as references. The appearance similarity and location distance can complement each other under crowded or occluded scenarios. Thus, we include both the location and appearance embedding into the track state. 

We add $M_{T-1}$ from the detection encoder at frame $T-1$ to $M_{T}$ at $T$ to obtain the joint memory~$\widetilde M_{T-1, T}$.
To aggregate the corresponding features for the track state prediction at frame $T$, we perform multi-scale deformable co-attention as Fig \ref{framework}~(c) at the center point $v_{T-1}$ of each track~(reference point) at frame $T-1$ based on $\widetilde M_{T-1, T}$. A linear projection is used to transform $v_{T-1}$ to the reference embedding, which is then used to generate the corresponding $L$ location offsets $\Delta v_{T-1} \in \mathbf{R}^{2L}$ and attention weights $a_{T-1} \in \mathbf{R}^{L}$ for the adaptive deformable co-attention. 
After aggregating the $\widetilde M_{T-1, T}$ features at $(v_{T-1}+\Delta v_{T-1})$ by the corresponding weights $a_{T-1}$, we apply two MLP heads to predict the center point location $\widetilde v_{T}$ at frame $T$ and the appearance embedding $\widetilde \varepsilon_T$ for each previous track one-to-one.

The tracks with $v_{T-1}$ outside the image region at frame $T-1$ are not considered as references to our RS module. We consider their clamped $v_{T-1}$ not the actual center points, and the extracted appearance for less than half body is unreliable to represent the object. Matching for RS is not applied between these tracks and the newly detected objects. The matching details will be described in the following contents. 

\noindent\textbf{Training.}
In this part, the goal of training is to make the RS module gain the ability to recognize the association between objects with the same identity from two adjacent frames. To this end, we only train the module when there exist object pairs that can be associated. Given the $i$-th track at frame $T-1$, we denote its center point location at frame $T-1$ as $v^i_{T-1}$, and the ground-truth center point location of the track at frame $T$ as $v^i_T$. Using $v^i_{T-1}$ as the reference points, we deploy the RS module to predict the center point location $\widetilde v^i_T$ and the appearance embedding $\widetilde \varepsilon^i_T$ for the $i$-th track at frame $T$. Then we calculate the L1 loss as $\mathcal{L}_{\operatorname {reg}}$ between $\widetilde v^i_T$ and the corresponding ground-truth $v^i_T$ to supervise the location prediction for the center point. 
Similar to the detection part, we construct the id loss $\mathcal{L}_{\operatorname{id}}$ for supervision on the appearance embedding prediction. The RS training loss $\mathcal{L}_{\operatorname {RS}}$ on $M$ references $Q_r$ is defined as:
\begin{equation}
\small \label{RS cost}
\hspace{-1.5mm}\mathcal{L}_{\operatorname {RS}}=\sum_{i=1}^{M}[\lambda_{reg}\mathcal{L}_{\operatorname {reg}}(v^i_T, \widetilde v^i_T)+ \lambda_{id}\mathcal{L}_{\operatorname {id}}(\alpha, \varphi'(\widetilde \varepsilon^i_T)]
\end{equation}

\noindent\textbf{Multi-layer RS.}
We construct the multi-layer RS for better performance, where each layer has the same structure. The aggregated features before MLP of each layer are utilized to update the reference embedding of the next layer. And $\widetilde \varepsilon_T$ is only predicted from the last layer. We apply the auxiliary losses~\cite{al2019character} after each layer for supervision.

\subsection{Matching and Track Updating} \label{match}
As Fig \ref{framework}~(d), after predicting the current state for each track, a corresponding matching strategy is designed to associate the newly detected objects to the existing tracks.

\noindent\textbf{Matching for RS.} \label{Matching for RS}
We assume a reliable state prediction from RS contains the appearance embedding consistent with the embedding of its corresponding track reference. Therefore, to assess the predicting reliability, we introduce Reference Consistency, which indicates the appearance similarity between the RS prediction and the reference.

Given the $i$-th track at frame $T-1$, we denote its appearance embedding as $\varepsilon_i$ and the center point location as $v^i_{T-1}$. Besides the $\widetilde v^i_T$ and $\widetilde \varepsilon^i_T$ that are predicted from the RS module, we apply the Kalman Filter~\cite{Kalman1960ANA} to predict the shape $(w^i_T, h^i_T)$ of the $i$-th track at frame T. In the detection process at frame $T$, for the $j$-th detected object, the predicted center point with the bounding box shape is denoted as $\hat l^j_T=(\hat v^j_T, \hat w^j_T, \hat h^j_T)$, and the appearance embedding is $\hat \varepsilon^j_T$. The RS matching cost between the $i$-th track and the $j$-th detected object consists of three parts: 1) the Euclidean distance between $\widetilde l^i_T=(\widetilde v^i_T, w^i_T, h^i_T)$ and $\hat l^j_T$; 2) the Reference Consistency between $\varepsilon_i$ and $\widetilde \varepsilon^i_T$; 3) the appearance similarity between $\widetilde \varepsilon^i_T$ and $\hat \varepsilon^j_T$. We utilize the cosine similarity to calculate the Reference Consistency and appearance similarity. The RS matching cost $\mathcal{C}_{RS}(i, j)$ between the $i$-th track and the $j$-th detected object is constructed as:
\begin{equation}
\small
\vspace{-2mm}
\hspace{-1.5mm}
\begin{split} \mathcal{C}_{RS}(i, j)= & ~\lambda_{\operatorname {emb}}\sqrt{\frac{\varepsilon_i \cdot \widetilde\varepsilon^i_T}{\|\varepsilon_i\|\|\widetilde\varepsilon^i_T\|} \cdot \frac{\widetilde\varepsilon^i_T \cdot \hat\varepsilon^j_T}{\|\widetilde\varepsilon^i_T\|\|\hat\varepsilon^j_T\|}}\\ & +(1-\lambda_{\operatorname {emb}})\|\widetilde l^i_T-\hat l^j_T\|
\end{split}
\end{equation}

\noindent\textbf{Matching pipeline.} The objects with detection scores over $\mu$ are preserved for further association. First, we calculate $\mathcal{C}_{RS}$ and find the bipartite matching with the least cost between the detections and the state predictions for previous tracks from RS. Only the matching that costs less than $\tau$ is valid. Then we associate the remaining detected objects with the remaining previous tracks that have the most location overlaps, where the overlaps should be greater than a threshold of $\theta$. After these two steps, the previous tracks that are not associated are denoted as the lost tracks, and the remaining detected objects are considered as the new unconfirmed tracks. We confirm a new track only if it is associated with a detected object at the next frame, otherwise we will remove the track.

\noindent\textbf{Track updating.} The associated previous tracks apply the corresponding matched objects to update their locations and appearance embedding. To re-identify the lost tracks, we use the Kalman Filter~\cite{Kalman1960ANA} to update their states when the consecutive lost frames are less than $\xi$. Otherwise, the tracks are terminated and become inactive. After that, all the active tracks, including the active lost tracks, are taken as RS references at the next frame. The effectiveness of the re-identifying strategy for track rebirth is verified in Sec~\ref{ablation for matching}.

\subsection{Joint Training} \label{sec:joint training}
A joint training strategy is designed for our proposed framework. Two adjacent frames are taken as the input, and we perform detection on the two frames, respectively. To make the training process more robust and consistent to the inference, we take the detected objects from the previous frame $T-1$ as references for RS at the latter frame $T$. The loss function of our joint training is defined as:
\begin{equation}
\small
\vspace{-2mm}
\mathcal{L}(T)=0.5(\mathcal{L}_{\operatorname {det}}(T-1) + \mathcal{L}_{\operatorname {det}}(T)) + \mathcal{L}_{\operatorname {RS}}(T-1, T),
\end{equation}
where $\mathcal{L}(T)$ and $\mathcal{L}(T-1, T)$ indicate the loss functions based on frame $T$ and the combination of frame $T-1$ with $T$, respectively. In the loss backpropagation process, we stop the gradient backpropagation of features from $T-1$.

\section{Experiments}
\subsection{Datasets and Metrics}
\noindent\textbf{Datasets.} We verify the effectiveness of our method on two datasets, \emph{i.e.}, MOT17~\cite{milan2016mot16} and MOT20~\cite{mot20}. MOT17 consists of $14$ sequences with heavy occlusions, which are split into $7$ sequences for training and $7$ sequences for testing. MOT20 consists of $8$ sequences depicting very crowded challenging scenes. 
  
\noindent\textbf{Metrics.} We apply the official evaluation metrics in MOT datasets. 1)~MOTA~\cite{Bernardin2008}: Summary of overall tracking accuracy in terms of false positives, false negatives and identity switches. 2)~IDF$1$~\cite{ristani2016performance}: The ratio of correctly identified detections over the average number of ground-truth and computed detections.
3)~MOTP~\cite{Bernardin2008}: The misalignment between the annotated and the predicted bounding boxes. 
4)~MT: Mostly tracked targets. The ratio of ground-truth trajectories that are covered by a track hypothesis for at least $80\%$ of their respective life span. 5)~ML: Mostly lost targets. The ratio of ground-truth trajectories that are covered by a track hypothesis for at most $20\%$ of their respective life span.
3)~HOTA~\cite{hotaLuiten_2020}: Geometric mean of detection accuracy and association accuracy. Averaged across localization thresholds. \label{metrics}
4)~IDS~\cite{li2009learning}: Number of identity switches. 

\subsection{Implementation Details} \label{implement}
We implement our detector following the setting of two-stage Deformable DETR with iterative bounding box refinement~\cite{zhu2020deformable}, utilizing $300$ object queries and a ResNet-50 backbone~\cite{he2016deep}. $d$ and $K$ in Fig \ref{framework} are $256$ and $64$ respectively. A $6$-layer RS module is applied, where $L$ is $12$ in total for all three scales. For training, we first pretrain our model on the CrowdHuman dataset with the MOT17 or MOT20 dataset and then finetuning on these two datasets while blocking CrowdHuman for appearance embedding training. 
For the CrowdHuman dataset without annotated identities, we give each object a unique identity to train the appearance embedding branch in a self-supervised way following~\cite{zhang2020fairmot}.
We first pretrain our model for $20$ epochs with a learning rate of $10^{-4}$, and then finetune for $25$ epochs with a learning rate of $10^{-4}$ decreased by $10$ times at the $20$th epoch. For both the training steps, the batch size is $12$, and the optimizer is AdamW~\cite{loshchilov2018decoupled}. We set the weight coefficients $\lambda_{\operatorname{reg}}, \lambda_{\operatorname{id}}$, $\lambda_{\operatorname{det}}$ and $\lambda_{\operatorname{RS}}$ in Sec~\ref{sec:method} to $5$, $0.5$, $1$ and $1$ respectively. For inference, the hyper-parameters $\mu$, $\tau$, $\theta$ and $\xi$ in the matching pipeline are set to $0.4$, $0.8$, $0.5$ and $30$ respectively. All experiments are conducted on the GTX $1080$Ti, CUDA $9.0$.

\begin{table*}[htb!]
  \begin{center}
  \small
  \resizebox{1.0\textwidth}{!}{%
  \vspace{1mm}  
  \begin{tabular}{c|c|cc|ccc}
\hline
Benchmark & Method & Detector & Extra Data & MOTA$\uparrow$ & IDF$1$$\uparrow$ & HOTA$\uparrow$\\
\hline
\multirow{15}{*}{MOT17} 
&SST~\cite{sun2019deep} & Faster R-CNN &\xmark & 52.4 & 49.5 & 39.3\\
&Tracktor+Ctdet~\cite{bergmann2019tracking} & Faster R-CNN &\xmark & 54.4 & 56.1 & -\\
&DeepSORT~\cite{wojke2017simple} & Faster R-CNN &\xmark & 60.3 & 61.2 & - \\
&TubeTK~\cite{pang2020tubetk} & 3D FCOS &\xmark & 63.0 & 58.6  & 48.0\\
&ChainedTracker~\cite{peng2020chained} & RetinaNet &\xmark & 66.6 & 57.4 & 49.0\\
&GSDT~\cite{wang2021joint} &CenterNet & 5D2 & 73.2 & 66.5 & 55.2\\
&FairMOT~\cite{zhang2020fairmot} & CenterNet &5D1 & 73.7 & 72.3 & 59.3\\ 
&CSTrack~\cite{liang2020rethinking} & YOLOv5 &5D1 & 74.9 & 72.6 & 59.3\\
&FUFET~\cite{fufetshan2020tracklets} &ResNet101-FPN & 5D1 & 76.2 & 68.0 & 57.9\\
&CorrTracker~\cite{corrtrackerwang2021multiple} & CenterNet & 5D1 & 76.5 & 73.6 &60.7\\
&Centertrack~\cite{zhou2020tracking} & CenterNet &CH & 67.8 & 64.7 & 52.2\\
&TraDeS~\cite{tradeswu2021track}& CenterNet & CH & 69.1 & 63.9 &52.7\\
&TransCenter~\cite{xutranscenter2021} &Deformable-DETR & CH & 73.2 & 62.2 & 54.5\\
&TransTrack~\cite{Sun2020TransTrackMT} & Deformable-DETR &CH & 75.2 & 63.5 & 54.1\\

&\cellcolor{mygray-bg} TR-MOT~(Ours)& \cellcolor{mygray-bg}Deformable-DETR& \cellcolor{mygray-bg}CH& \cellcolor{mygray-bg}76.5&  \cellcolor{mygray-bg}72.6 &  \cellcolor{mygray-bg}59.7\\
\hline
\multirow{6}{*}{MOT20}  
&FairMOT~\cite{zhang2020fairmot} &CenterNet & 5D1 & 61.8 & 67.3 &54.6\\
&CSTrack~\cite{liang2020rethinking} &YOLOv5 & 5D1 & 66.6 & 68.6 &54.0\\
&GSDT~\cite{wang2021joint} &CenterNet & 5D2 & 67.1 & 67.5 & 53.6\\
&TransCenter~\cite{xutranscenter2021} &Deformable-DETR & CH & 58.3 & 46.8 & 44.3\\
&TransTrack~\cite{Sun2020TransTrackMT} &Deformable-DETR & CH & 64.5 & 59.2 & 48.5\\

&\cellcolor{mygray-bg} TR-MOT~(Ours)& \cellcolor{mygray-bg}Deformable-DETR& \cellcolor{mygray-bg}CH& \cellcolor{mygray-bg}67.1&  \cellcolor{mygray-bg}59.1 &  \cellcolor{mygray-bg}50.4\\ \hline
\end{tabular}}
  \end{center}
  \vspace{-3mm}
    \caption{Performance comparison on the MOT17 and MOT20 test set with private detector. 5D1 indicates using 5 datasets, including CH~\cite{shao2018crowdhuman}, Caltech Pedestrian~\cite{caltechdollar2009pedestrian}, CityPersons~\cite{citypersonzhang2017citypersons}, CUHK-SYS~\cite{cuhkxiao2017joint}, and PRW~\cite{prwzheng2017person}, 5D2 is 5D1 replacing CH by ETH~\cite{ethess2008mobile}}
 \label{tb:mot17&20}
 \end{table*}

\subsection{Main Results}
We conduct experiments on two benchmarks to verify our effectiveness. In Table  \ref{tb:mot17&20}, we compare our performance with the reported results of previous online methods.
On MOT17, we achieve $76.5\%$ on MOTA, $72.6\%$ on IDF$1$ and $59.7\%$ on HOTA, which outperform the state-of-the-art. Moreover, comparing to the methods ~\cite{Sun2020TransTrackMT, xutranscenter2021}, which also adopt a Deformable DETR detector, our TR-MOT largely outperforms on IDF$1$ and HOTA by $9.1\%$ and $5.2\%$ respectively, and also outperforms on detection related metrics MOTA$1$ by $1.3\%$.
On MOT20, we achieve $67.1\%$ on MOTA, $59.1\%$ on IDF$1$ and $50.4\%$ on HOTA. In the extremely crowded MOT20 datasets, the appearance of each human is of relatively low quality and not easily to be differentiated. Some methods~\cite{zhang2020fairmot, liang2020rethinking, wang2021joint} use additional ReID datasets like~\cite{cuhkxiao2017joint, prwzheng2017person, ethess2008mobile} to train the appearance embedding branch with high-quality human appearances and clear identities, to further improve their association ability. Comparing to these methods on MOT20, our method without utilizing any ReID dataset does not have superiority on association. However, we outperform other methods which use the the same datasets and detector as ours for training. The main results indicate that our performance boosts from our well designed MOT pipeline with the associating module RS, rather than only from the detector. We achieve competitive association ability while maintaining the high detection performance.

\begin{table*}[t]
\centering
\vspace{1mm}
\subfloat[\textbf{Ablations on Sequences.} The value in the table indicates two important metrics IDF$1\uparrow$~/~MOTA$\uparrow$ on the validation sequences. $*$ denotes the sequence shot by an apparently moving camera, and $\dag$ means the validation sequence is crowded and with a heavy occlusion.\label{tab:ablation:seq}]{
\tablestyle{2.2pt}{1.05}\begin{tabular}{c|x{52}x{52}x{52}x{52}x{52}x{52}x{52}}
 {Variant} &  {Video-02$\dag$} &  {Video-04}  &  {Video-05$*\dag$} & {Video-09} &  {Video-10$*$} &  {Video-11$*\dag$} &  {Video-13$*$} \\
\shline
 \emph{\scriptsize {Baseline Model, w/o RS}} &{42.8~/~42.1} &  {82.3~/~85.7}& {59.3~/~65.7}& {74.8~/~78.8}& {65.6~/~61.2}&  {45.9~/~68.1}& {82.0~/~65.2} \\
\hline
\emph{\scriptsize {+ $3\times$ RS w/o appearance}} & {43.8~/~44.8} & {82.2~/~86.7}& {75.3~/~69.3}& {74.1~/~79.1}& {60.7~/~61.0}& {61.8~/~69.4}& {71.7~/~61.8} \\
\emph{\scriptsize {+ $3\times$ RS w/ appearance}} & {46.5~/~46.0} &  {84.1~/~85.7}& {71.0~/~67.2}& {70.7~/~78.9}& {71.0~/~63.3}&  {76.7~/~68.6}& {79.8~/~62.6} \\
\hline
\emph{\scriptsize {+ $3\times$ RS w/o appearance $\Delta$}} & +2.3\%/+6.4\% & -0.1\%/+1.2\% & +27\%/+5.5\% & -0.9\%/+0.4\% & -7.5\%/-0.3\% & +35\%/+1.9\% & -12\%/-5.2\% \\
\emph{\scriptsize {+ $3\times$ RS w/ appearance $\Delta$}} & +8.6\%/+9.3\%& +2.2\%/+0.0\% & +20\%/+2.3\% & -5.5\%/+0.1\% & +8.2\%/+3.4\% & +67\%/+0.7\% & -2.7\%/-4.0\% \\
\end{tabular}}\hspace{3mm}
\subfloat[\textbf{Component Analysis:} Results of the variants with various components, \ie. The detection threshold $\mu =0.4$, and RS matching cost threshold $\tau =1.5$. \label{tab:ablation:component}]{
\tablestyle{2.2pt}{1.05}\begin{tabular}{c|c|c|x{32}x{32}x{32}x{32}x{32}|x{40}}
\scriptsize {Variant} & \scriptsize {RS Layers} & \scriptsize State Prediction &IDF$1$$\uparrow$   &MOTA$\uparrow$ &IDS$\downarrow$ &MT$\uparrow$ &ML$\downarrow$ &\scriptsize{\#Parameters} \\
\shline
 \emph{Baseline, w/o RS} &- & \scriptsize Kalman Filter &69.1  &70.8   &382    &161 &50 &41.0~M\\
\hline
\emph{+ RS w/o appearance} & $3\times$ & \scriptsize Reference Search &70.7    &72.0   &607    &\bd{164}   &51  &44.7~M\\
\hline
&$1\times$ & \scriptsize Reference Search & 73.5& \bd{72.9}&  321&  152& 51& 42.6~M\\
\emph{+ RS w/ appearance}&$3\times$ & \scriptsize Reference Search &74.2 &71.9   &287    &160    &\bd{48} &44.9~M\\
&$6\times$ &\scriptsize Reference Search & \bd{76.0} &72.5 &\bd{233} &162 &51 &48.3~M\\
\end{tabular}}\\ \hspace{3mm}
\vspace{-1.5mm}
\subfloat[\textbf{Threshold of $\mathcal{C}_{RS}$.} $\mu =0.4$.\label{ablation:tau}]{
\tablestyle{2.5pt}{1.05}\begin{tabular}{c|x{28}x{28}x{28}}
 \scriptsize {$\tau$} &IDF$1$ $\uparrow$   & MOTA$\uparrow$ & IDS$\downarrow$\\
\shline
 \scriptsize \emph{0.5} & 73.6&  70.8& 656\\
 \scriptsize \emph{0.8} & \bd{76.5}&  72.4& 278\\
 \scriptsize \emph{1.0} & 76.3&  72.4& 266\\
 \scriptsize \emph{1.2} & 76.1&  72.4& 245\\
 \scriptsize \emph{1.5} & 76.0&  \bd{72.5}& \bd{233}\\
 \scriptsize \emph{1.8} & 75.0&  72.4& 238\\
\end{tabular}}
\subfloat[\textbf{Matching Pipeline.} $\tau =0.8$, $\mu =0.4$.\label{ablation:matching pipeline}]{
\tablestyle{4.8pt}{1.05}\begin{tabular}{c|x{28}x{28}x{28}}
\scriptsize {Matching Pipeline} &IDF$1$ $\uparrow$  &MOTA$\uparrow$ &IDS$\downarrow$\\
\shline
 \scriptsize \emph{IoU + confirm} & 63.4& 69.1& 910\\
 \scriptsize \emph{JDE+IoU+confirm} & 69.1& 70.8& 382\\ \hline
 \scriptsize \emph{RS loc. matching} & 61.0& 69.3& 1056\\
 \scriptsize \emph{RS loc.+RS App.} & 63.0&  69.0& 1131\\
 \scriptsize \emph{Full RS~(RS w/ lost ref)} & 71.3& 68.3& 1681\\
 \scriptsize \emph{Full RS + confirm} & 74.9& 70.8& 616\\ \hline
 \scriptsize \emph{(RS w/o lost ref)+IoU+confirm} & 65.2& 70.2& 575\\
 \scriptsize \emph{Full RS+IoU+confirm} & \bd{76.5}& \bd{72.4}& \bd{278}\\
\end{tabular}}
\vspace{-1.5mm}
\subfloat[\textbf{Reference Consistency Strategy.} $\tau =0.8$, $\mu =0.4$.\label{ablation:Reference Consistency}]{
\tablestyle{2.5pt}{1.05}\begin{tabular}{c|x{28}x{28}x{28}}
\scriptsize {Strategy} &IDF$1$ $\uparrow$ &MOTA$\uparrow$ &IDS$\downarrow$\\
\shline
 \scriptsize \emph{w/o} & 75.0&  72.1& 375\\
 \scriptsize \emph{average} & 75.7& 72.3& 299\\
 \scriptsize \emph{sqrt}& \bd{76.5}& \bd{72.4}& \bd{278} \\
 \multicolumn{4}{c}{~}\\
 \multicolumn{4}{c}{~}\\
\end{tabular}}
\caption{Ablation studies of our proposed model on the MOT17 validation set.}
\label{tab:ablations}
\end{table*}

\subsection{Ablation Studies}
For ablation studies, we first pretrain our model using CrowdHuman and finetune on the first half of MOT17 training set~\cite{zhang2020fairmot}, following other settings introduced in the implementation details. Then we evaluate the trained models on the other half of MOT17 training set. We focus on three significant MOT metrics: MOTA, IDF$1$ and IDS. According to previous metrics definition, MOTA, IDS and IDF$1$ mainly indicate the detection performance, association stability, and association reliability, respectively. Considering most variants we implement mainly differ in the association process, we give priority to the metric IDF$1$. 

\noindent\textbf{Baseline model.} For a fair comparison to verify the effectiveness of our proposed method, we follow JDE~\cite{wang2019towards} to implement a variant as our baseline model and replace the detector with the same Deformable DETR detector as ours. We simultaneously output detection results and the corresponding appearance embedding from the last decoder layer, then associate the detected objects with previous tracks based on appearance affinity and location distance.

\noindent\textbf{Proposed framework with RS.}
We implement our proposed pipeline with a $3$-layer RS module~(\emph{{+ $3\times$ RS w/ appearance}}) and also a variant without learning the appearance embedding in the $3$-layer RS module~(\emph{{+ $3\times$ RS w/o appearance}}). As shown in Table \ref{tab:ablation:seq}, we compare the results of these two variants with the baseline model on the validation sequences, respectively. Both two variants with our proposed RS module have comparatively superior performance on the sequence with a moving camera or under heavy occlusion. Especially, on the Video-$05$ and Video-$11$ which are with both heavy occlusion and apparent camera motion, \emph{{+ $3\times$ RS w/ appearance}} outperforms 20\% and 67\% relatively over the baseline model on the id association relevant metric IDF$1$, respectively. We consider it is because the baseline model does not leverage any visual temporal information like our RS module, which makes it not easier to handle unexpected erratic motions. Moreover, Table \ref{tab:ablation:component} shows the variants with RS yield $1.6\%$ - $6.9\%$ and $1.1\%$ - $2.1\%$ improvements on IDF$1$ and MOTA respectively, comparing to the baseline model.

\noindent\textbf{Importance of RS appearance.}
According to Table \ref{tab:ablation:seq}, RS with appearance learning~(\emph{{+ $3\times$ RS w/ appearance}}) outperforms the variant without appearance~(\emph{{+ $3\times$ RS w/o appearance}}) in most sequences that are with heavy occlusion or camera motion. Table \ref{tab:ablation:component} shows \emph{{+ $3\times$ RS w/ appearance}} gains a $3.5\%$ improvement over \emph{{+ $3\times$ RS w/o appearance}} on IDF$1$, and the IDS is dropped by $52.7$\% relatively over \emph{{+ $3\times$ RS w/o appearance}} when RS appearance learning is involved. We consider the appearance embedding obtained from our RS module is more representative by leveraging the temporal information. Since the RS module mainly contributes to the association, the detection relevant metric MOTA is less affected. With a relative similar detection performance, the apparent improvement on IDF$1$ and dropping on IDS can verify the effectiveness of the RS appearance for a more stably accurate association. 

\noindent\textbf{Number of Reference Search layers.}
We construct a multi-layer RS structure in our TR-MOT. As shown in Table \ref{tab:ablation:component}, row $3$ - $5$,  IDF$1$ is improved by $2.5\%$ from $1$ layer to $6$ layers, and IDS is dropped from $321$ to $233$ in a stable way. The improvements of IDF$1$ and dropping of IDS along with the close detection performance can verify our RS module with more layers can benefit the association performance.

\noindent\textbf{Matching cost for RS.}
During the matching for RS, we calculate the matching cost $\mathcal{C}_{RS}$ through Sec \ref{Matching for RS}. Based on our optimal model structure with a $6$-layer RS module~(\emph{{+ $6\times$ RS w/ appearance}}), we set the maximum threshold $\tau$ of $\mathcal{C}_{RS}$ from $0.5$ to $1.8$ and keep all other settings following \ref{implement} to perform the sensitivity analysis on $\tau$. As shown in Table \ref{ablation:tau}, IDF$1$ improves by $2.9\%$ when $\tau$ changes from $0.5$ to $0.8$, drops by only $0.5\%$ when $\tau$ increases from $0.8$ to $1.5$, and drops by $1.0\%$ as $\tau$ changes from $1.5$ to $1.8$. Moreover, the detection relevant metric MOTA keeps almost unchanged when $\tau$ is from $0.8$ to $1.8$. Table \ref{ablation:tau} shows the association performance has  consistent effectiveness when $\tau$ is set from $0.8$ to $1.5$, and is best when $\tau=0.8$. When $\tau$ is smaller, the matching for RS will be more strict and thus results in more unreliable new tracks to be unconfirmed and removed. Therefore, the detection recall will drop and degrade the detection relevant metric MOTA, which explains the $1.6\%$ dropping on MOTA when $\tau$ decreases from $0.8$ to $0.5$.

\noindent\textbf{Matching for RS.}
Rows $3$ - $5$ in Table \ref{ablation:matching pipeline} show the ablations on the matching for RS. When lost tracks are excluded from the references and only RS matching is utilized, involving RS appearance into the RS matching cost~(\emph{RS loc.+RS App.}) can improve IDF$1$ by $2\%$ with very close MOTA comparing to only considering the updated track location from RS~(\emph{RS loc. matching}). It verifies the association effectiveness of jointly considering RS appearance and location overlaps in the matching process for RS. 

\noindent\textbf{Strategy for Reference Consistency.}
According to Sec \ref{Matching for RS}, we introduce Reference Consistency to assess the reliability of RS predictions. As Table \ref{ablation:Reference Consistency}, we implement two strategic variants for involving the Reference Consistency to RS matching. When calculating the RS matching cost, if we take the average value of the Reference Consistency and appearance similarity~(\emph{average}), IDF$1$ and MOTA will drop by $0.8\%$ and $0.1\%$, respectively, comparing to taking the sqrt value. If we exclude the Reference Consistency, IDF$1$ and MOTA will further drop by $0.7\%$ and $0.2\%$ respectively comparing to \emph{average}. Therefore, Table \ref{ablation:Reference Consistency} verifies the contribution of our proposed Reference Consistency.

\noindent\textbf{Rebirth in RS.} \label{ablation for matching}
If we add the lost tracks into the track references~(\emph{Full RS~(RS w/ lost ref)}) based on \emph{RS loc. Matching}, IDF$1$ will be improved largely by $8.3\%$, although MOTA drops by $0.7\%$. However, \emph{Full RS+IoU+confirm} improves IDF$1$ by $11.3\%$ over \emph{(RS w/o lost ref)+IoU+confirm} while MOTA increases by $2.2\%$ and IDS drops by $297$. Without the \emph{confirm} process, more lost tracks in low quality as references will impair the RS matching and increase the IDS, which degrades the metric MOTA and IDF1. When we involve \emph{confirm} based on \emph{Full RS~(RS w/ lost ref)}, \emph{Full RS + confirm} will improve IDF$1$ and MOTA by $3.6\%$ and $2.5\%$, respectively. These ablations indicate that considering the lost tracks in RS for track rebirth is necessary, and also the performance benefits from the \emph{confirm} process.

\noindent\textbf{Matching pipeline.}
For the full matching pipeline~(\emph{Full RS+IoU+confirm}), we perform matching for RS~(\emph{full RS}), and deploy matching for the location overlaps~(\emph{IoU}) to associate the remaining detected objects and tracks. Moreover, the unconfirmed tracks that are not associated to new detections will be removed~(\emph{confirm}). To verify the effectiveness of our designed matching pipeline with RS, we implement several variants while keeping the hyperparameters. If we exclude \emph{full RS} for matching~(\emph{IoU+confirm}), IDF$1$ drops greatly by $13.1\%$ and MOTA also drops by $3.3\%$. If we replace \emph{full RS} to the matching strategy based on the appearance embedding and detection results from the detector as JDE~\cite{wang2019towards}~(\emph{JDE}), IDF$1$ drops by $7.4\%$ and MOTA drops by $1.6\%$. These two ablations verify the large contribution to the association is from our \emph{full RS} in the matching pipeline. Overall, our designed matching pipeline~(\emph{Full RS+IoU+confirm}) can achieve the best result.

\subsection{Visualization}
\begin{figure}[ht]
  \centering
  \includegraphics[width=1.0\linewidth]{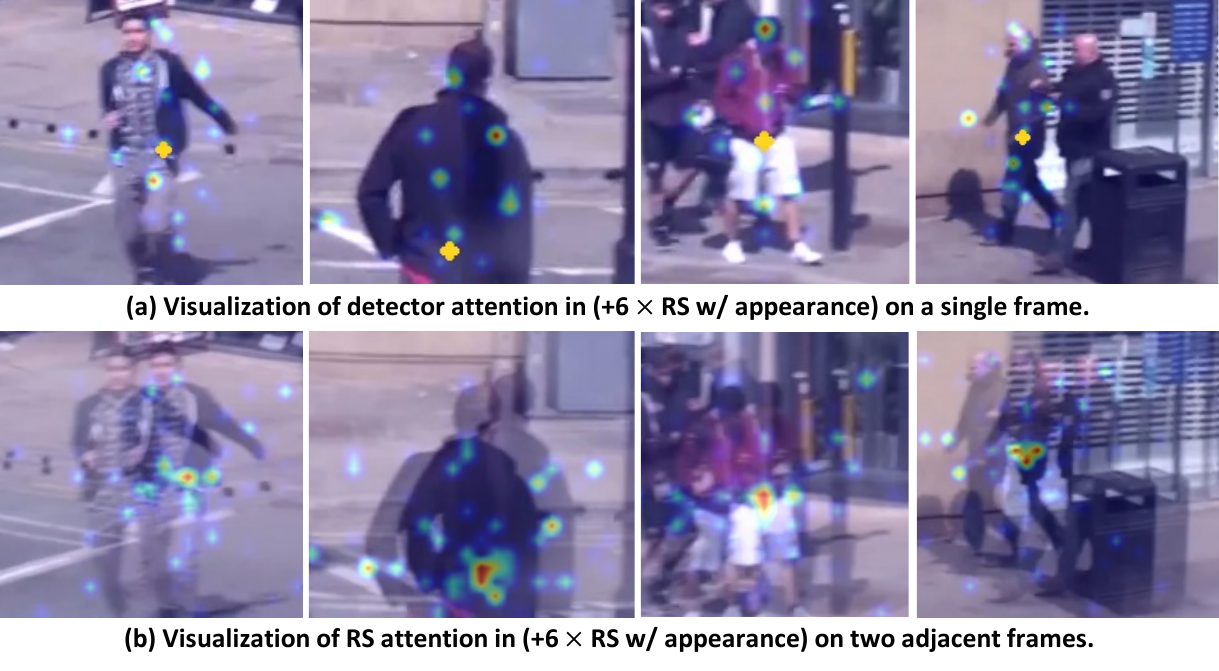}
  \caption{Visualization for the detector attention and RS attention. The sub-figures in (a) and (b) have a one-to-one correspondence. The yellow cross in sub-figure (a) indicates the reference location of RS in the corresponding sub-figure (b).}\vspace{-5mm}
\label{fig:visualization}
\end{figure}

As shown in Fig \ref{fig:visualization}, we visualize the attention of our detector on one frame~(a) and RS attention on two adjacent frames~(b) in (\emph{{+ $6\times$ RS w/ appearance}}). From the figure, the detector focuses more on the keypoints of the human body to predict an accurate human location. Differently, our RS module concerns features of both humans that are associated with each other, especially the aligned body features from adjacent frames near the reference locations. Some contexts are also taken into account by the RS module for a more reliable association. The visualization indicates our RS module can cognize the appearance feature alignment for each track between adjacent frames. It also explains why the predicted appearance embedding from RS is more representative and thus benefits the association process.

\section{Conclusion}
In this paper, we propose an effective online MOT framework TR-MOT. Specifically, we design a Reference Search module for more reliable association, which takes previous tracks as references to predict the current state for each track reference, based on the visual temporal information. Compatibly, We propose an effective matching pipeline to associate new detections with the predicted track states and then update the existing tracks. Moreover, we introduce a joint training strategy for the end-to-end training. Our method achieves competitive performance on the MOT dataset.

{\small
\bibliographystyle{ieee_fullname}
\bibliography{egbib}

\begin{thebibliography}{10}\itemsep=-1pt

\bibitem{al2019character}
Rami Al-Rfou, Dokook Choe, Noah Constant, Mandy Guo, and Llion Jones.
\newblock Character-level language modeling with deeper self-attention.
\newblock In {\em AAAI}, 2019.

\bibitem{DBLP:journals/corr/abs-1801-09646}
David{-}Alexandre Beaupr{\'{e}}, Guillaume{-}Alexandre Bilodeau, and Nicolas
  Saunier.
\newblock Improving multiple object tracking with optical flow and edge
  preprocessing.
\newblock {\em CoRR}, abs/1801.09646, 2018.

\bibitem{bergmann2019tracking}
Philipp Bergmann, Tim Meinhardt, and Laura Leal-Taixe.
\newblock Tracking without bells and whistles.
\newblock In {\em ICCV}, 2019.

\bibitem{Bernardin2008}
Keni Bernardin and Rainer Stiefelhagen.
\newblock Evaluating multiple object tracking performance: the clear mot
  metrics.
\newblock {\em EURASIP Journal on Image and Video Processing}, 2008.

\bibitem{braso2020learning}
Guillem Brasó and Laura Leal-Taixé.
\newblock Learning a neural solver for multiple object tracking.
\newblock In {\em CVPR}, 2020.

\bibitem{carion2020endtoend}
Nicolas Carion, Francisco Massa, Gabriel Synnaeve, Nicolas Usunier, Alexander
  Kirillov, and Sergey Zagoruyko.
\newblock End-to-end object detection with transformers.
\newblock In {\em ECCV}, 2020.

\bibitem{pengtransmot2021}
Peng Chu, Jiang Wang, Quanzeng You, Haibin Ling, and Zicheng Liu.
\newblock Transmot: Spatial-temporal graph transformer for multiple object
  tracking.
\newblock {\em CoRR}, abs/2104.00194, 2021.

\bibitem{mot20}
Patrick Dendorfer, Hamid Rezatofighi, Anton Milan, Javen Shi, Daniel Cremers,
  Ian~D. Reid, Stefan Roth, Konrad Schindler, and Laura Leal{-}Taix{\'{e}}.
\newblock {MOT20:} {A} benchmark for multi object tracking in crowded scenes.
\newblock {\em CoRR}, abs/2003.09003, 2020.

\bibitem{caltechdollar2009pedestrian}
Piotr Doll{\'a}r, Christian Wojek, Bernt Schiele, and Pietro Perona.
\newblock Pedestrian detection: A benchmark.
\newblock In {\em CVPR}, pages 304--311. IEEE, 2009.

\bibitem{ethess2008mobile}
Andreas Ess, Bastian Leibe, Konrad Schindler, and Luc Van~Gool.
\newblock A mobile vision system for robust multi-person tracking.
\newblock In {\em CVPR}, pages 1--8. IEEE, 2008.

\bibitem{he2016deep}
Kaiming He, Xiangyu Zhang, Shaoqing Ren, and Jian Sun.
\newblock Deep residual learning for image recognition.
\newblock In {\em CVPR}, 2016.

\bibitem{Kalman1960ANA}
R. Kalman.
\newblock A new approach to linear filtering and prediction problems,
  transaction of the asme~journal of basic.
\newblock 1960.

\bibitem{hungarian}
H.~W. Kuhn.
\newblock The hungarian method for the assignment problem.
\newblock In {\em Naval Research Logistics Quarterly}, 1955.

\bibitem{li2009learning}
Yuan Li, Chang Huang, and Ram Nevatia.
\newblock Learning to associate: Hybridboosted multi-target tracker for crowded
  scene.
\newblock In {\em CVPR}, 2009.

\bibitem{liang2020rethinking}
Chao Liang, Zhipeng Zhang, Yi Lu, Xue Zhou, Bing Li, Xiyong Ye, and Jianxiao
  Zou.
\newblock Rethinking the competition between detection and reid in multi-object
  tracking.
\newblock {\em arXiv preprint arXiv:2010.12138}, 2020.

\bibitem{lin2017focal}
Tsung-Yi Lin, Priya Goyal, Ross Girshick, Kaiming He, and Piotr Doll{\'a}r.
\newblock Focal loss for dense object detection.
\newblock In {\em ICCV}, 2017.

\bibitem{loshchilov2018decoupled}
Ilya Loshchilov and Frank Hutter.
\newblock Decoupled weight decay regularization.
\newblock In {\em ICLR}, 2018.

\bibitem{hotaLuiten_2020}
Jonathon Luiten, Aljosa Osep, Patrick Dendorfer, Philip H.~S. Torr, Andreas
  Geiger, Laura Leal{-}Taix{\'{e}}, and Bastian Leibe.
\newblock {HOTA:} {A} higher order metric for evaluating multi-object tracking.
\newblock {\em International Journal of Computer Vision}, 129(2):548–578, Oct
  2020.

\bibitem{mahmoudi2019multi}
Nima Mahmoudi, Seyed~Mohammad Ahadi, and Mohammad Rahmati.
\newblock Multi-target tracking using cnn-based features: Cnnmtt.
\newblock {\em Multimedia Tools and Applications}, 2019.

\bibitem{meinhardt2021trackformer}
Tim Meinhardt, Alexander Kirillov, Laura Leal-Taixe, and Christoph
  Feichtenhofer.
\newblock Trackformer: Multi-object tracking with transformers.
\newblock {\em arXiv preprint arXiv:2101.02702}, 2021.

\bibitem{milan2016mot16}
Anton Milan, Laura Leal-Taix{\'e}, Ian Reid, Stefan Roth, and Konrad Schindler.
\newblock Mot16: A benchmark for multi-object tracking.
\newblock {\em arXiv preprint arXiv:1603.00831}, 2016.

\bibitem{pang2020tubetk}
Bo Pang, Yizhuo Li, Yifan Zhang, Muchen Li, and Cewu Lu.
\newblock Tubetk: Adopting tubes to track multi-object in a one-step training
  model.
\newblock In {\em CVPR}, 2020.

\bibitem{peng2020chained}
Jinlong Peng, Changan Wang, Fangbin Wan, Yang Wu, Yabiao Wang, Ying Tai,
  Chengjie Wang, Jilin Li, Feiyue Huang, and Yanwei Fu.
\newblock Chained-tracker: Chaining paired attentive regression results for
  end-to-end joint multiple-object detection and tracking.
\newblock In {\em ECCV}, 2020.

\bibitem{ristani2016performance}
Ergys Ristani, Francesco Solera, Roger Zou, Rita Cucchiara, and Carlo Tomasi.
\newblock Performance measures and a data set for multi-target, multi-camera
  tracking.
\newblock In {\em ECCV}, 2016.

\bibitem{fufetshan2020tracklets}
Chaobing Shan, Chunbo Wei, Bing Deng, Jianqiang Huang, Xian-Sheng Hua,
  Xiaoliang Cheng, and Kewei Liang.
\newblock Tracklets predicting based adaptive graph tracking, 2020.

\bibitem{shao2018crowdhuman}
Shuai Shao, Zijian Zhao, Boxun Li, Tete Xiao, Gang Yu, Xiangyu Zhang, and Jian
  Sun.
\newblock Crowdhuman: A benchmark for detecting human in a crowd.
\newblock {\em arXiv preprint arXiv:1805.00123}, 2018.

\bibitem{Sun2020TransTrackMT}
Peize Sun, Yi Jiang, Rufeng Zhang, Enze Xie, Jinkun Cao, Xinting Hu, Tao Kong,
  Zehuan Yuan, Changhu Wang, and Ping Luo.
\newblock Transtrack: Multiple-object tracking with transformer.
\newblock {\em arXiv preprint arXiv:2012.15460}, 2020.

\bibitem{sun2019deep}
ShiJie Sun, Naveed Akhtar, HuanSheng Song, Ajmal Mian, and Mubarak Shah.
\newblock Deep affinity network for multiple object tracking.
\newblock {\em TPAMI}, 2019.

\bibitem{corrtrackerwang2021multiple}
Qiang Wang, Yun Zheng, Pan Pan, and Yinghui Xu.
\newblock Multiple object tracking with correlation learning, 2021.

\bibitem{wang2021joint}
Yongxin Wang, Kris Kitani, and Xinshuo Weng.
\newblock Joint object detection and multi-object tracking with graph neural
  networks, 2021.

\bibitem{wang2019towards}
Zhongdao Wang, Liang Zheng, Yixuan Liu, and Shengjin Wang.
\newblock Towards real-time multi-object tracking.
\newblock {\em arXiv preprint arXiv:1909.12605}, 2019.

\bibitem{wojke2017simple}
Nicolai Wojke, Alex Bewley, and Dietrich Paulus.
\newblock Simple online and realtime tracking with a deep association metric.
\newblock In {\em ICIP}, 2017.

\bibitem{tradeswu2021track}
Jialian Wu, Jiale Cao, Liangchen Song, Yu Wang, Ming Yang, and Junsong Yuan.
\newblock Track to detect and segment: An online multi-object tracker, 2021.

\bibitem{cuhkxiao2017joint}
Tong Xiao, Shuang Li, Bochao Wang, Liang Lin, and Xiaogang Wang.
\newblock Joint detection and identification feature learning for person
  search.
\newblock In {\em CVPR}, pages 3415--3424, 2017.

\bibitem{xutranscenter2021}
Yihong Xu, Yutong Ban, Guillaume Delorme, Chuang Gan, Daniela Rus, and Xavier
  Alameda{-}Pineda.
\newblock Transcenter: Transformers with dense queries for multiple-object
  tracking.
\newblock {\em CoRR}, abs/2103.15145, 2021.

\bibitem{yu2016poi}
Fengwei Yu, Wenbo Li, Quanquan Li, Yu Liu, Xiaohua Shi, and Junjie Yan.
\newblock Poi: Multiple object tracking with high performance detection and
  appearance feature.
\newblock In {\em ECCV}, 2016.

\bibitem{zeng2021motr}
Fangao Zeng, Bin Dong, Tiancai Wang, Cheng Chen, Xiangyu Zhang, and Yichen Wei.
\newblock Motr: End-to-end multiple-object tracking with transformer, 2021.

\bibitem{citypersonzhang2017citypersons}
Shanshan Zhang, Rodrigo Benenson, and Bernt Schiele.
\newblock Citypersons: A diverse dataset for pedestrian detection.
\newblock In {\em CVPR}, pages 3213--3221, 2017.

\bibitem{zhang2020fairmot}
Yifu Zhang, Chunyu Wang, Xinggang Wang, Wenjun Zeng, and Wenyu Liu.
\newblock Fairmot: On the fairness of detection and re-identification in
  multiple object tracking.
\newblock {\em arXiv e-prints}, pages arXiv--2004, 2020.

\bibitem{prwzheng2017person}
Liang Zheng, Hengheng Zhang, Shaoyan Sun, Manmohan Chandraker, Yi Yang, and Qi
  Tian.
\newblock Person re-identification in the wild.
\newblock In {\em CVPR}, pages 1367--1376, 2017.

\bibitem{zhou2020tracking}
Xingyi Zhou, Vladlen Koltun, and Philipp Kr{\"a}henb{\"u}hl.
\newblock Tracking objects as points.
\newblock In {\em ECCV}, 2020.

\bibitem{zhou2018online}
Zongwei Zhou, Junliang Xing, Mengdan Zhang, and Weiming Hu.
\newblock Online multi-target tracking with tensor-based high-order graph
  matching.
\newblock In {\em ICPR}, 2018.

\bibitem{zhu2020deformable}
Xizhou Zhu, Weijie Su, Lewei Lu, Bin Li, Xiaogang Wang, and Jifeng Dai.
\newblock Deformable detr: Deformable transformers for end-to-end object
  detection.
\newblock In {\em ICLR}, 2020.

\bibitem{zhu2017flow}
Xizhou Zhu, Yujie Wang, Jifeng Dai, Lu Yuan, and Yichen Wei.
\newblock Flow-guided feature aggregation for video object detection.
\newblock In {\em ICCV}, 2017.

\end{thebibliography}
}

\end{document}